# Laplacian regularized low rank subspace clustering

**Song Yu[a], Wu Yiquan[a, b, c, d]**

[a] Nanjing University of Aeronautics and Astronautics, School of Electronic and Information Engineering, Jiangjun Avenue No. 29, Nanjing, China, 211106

[b] Key Laboratory of Port, Waterway and Sedimentation Engineering of the Ministry of Transport, Nanjing Hydraulic Research Institute, Guangzhou Street No. 223, Nanjing, China, 210029

[c] State Key Laboratory of Urban Water Resource and Environment, Nangang District, Huanghe Street No. 73, Harbin, China, 150090

[d] The Key Laboratory of Rivers and Lakes Governance and Flood Protection of Yangtse River Water Conservancy Committee, Huangpu Street No. 23, Wuhan, China, 430010

**Corresponding author:** Song Yu, e-mail: 519559374@qq.com, postal address: Nanjing University of Aeronautics and Astronautics, School of Electronic and Information Engineering, Jiangjun Avenue No. 29, Nanjing, China, 211106, telephone: 86 13952027101

**Abstract:** The problem of fitting a union of subspaces to a collection of data points drawn from multiple subspaces is considered in this paper. In the traditional low rank representation model, the dictionary used to represent the data points is chosen as the data points themselves and thus the dictionary is corrupted with noise. This problem is solved in the low rank subspace clustering model which decomposes the corrupted data matrix as the sum of a clean and self-expressive dictionary plus a matrix of noise and gross errors. Also, the clustering results of the low rank representation model can be enhanced by using a graph of data similarity. This model is called Laplacian regularized low rank representation model with a graph regularization term added to the objective function. Inspired from the above two ideas, in this paper a Laplacian regularized low rank subspace clustering model is proposed. This model uses a clean dictionary to represent the data points and a graph regularization term is also incorporated in the objective function. Experimental results show that, compared with the traditional low rank representation model, low rank subspace clustering model and several other state-of-the-art subspace clustering models, the model proposed in this paper can get better subspace clustering results with lower clustering error.

## 1. Introduction

Specialists and practitioners in the community of image processing, pattern recognition and machine learning often have to deal with very high dimensional data samples. The high-dimensionality of the data samples often brings several difficulties to the processing algorithms. First, the computational complexity and memory requirements of the processing algorithms will be very high. Second, as compared to the ambient space dimension, the number of samples is insufficient. These difficulties are commonly referred to as "the curse of dimensionality" [1]. Fortunately, these high dimensional data samples often lie in some low dimensional structure embedded in the high dimensional ambient space. These low dimensional structures are modeled by linear subspaces. If the data samples are drawn from one class, single linear subspace models can be used to model the data samples. The most widely used single linear subspace model is principal component analysis (PCA) [2]. In PCA, each data sample is represented as a linear combination of basis vectors of a linear subspace with much lower dimension compared to the ambient space. When the data samples are corrupted by gross errors, using PCA can cause large approximation errors. In this case, robust principle component analysis (RPCA) [3] is proposed. RPCA tries to decompose the data matrix as the superposition of a low rank matrix and a sparse matrix. Another well-known single linear subspace model is nonnegative matrix factorization (NMF) [4]. In NMF, the basis vectors used to represent each data sample are from a positive linear subspace. In most data representation and classification scenario, data samples are drawn from multiple classes. In such as a case, PCA and NMF cannot model these data samples well. Since the data samples from each class can be modeled by a linear subspace, the whole data samples can be modeled by a union of linear subspaces. Subspace clustering models which aim at clustering and representing data samples drawn from multiple classes are thus proposed. Subspace clustering has found wide applications in computer vision [5], [6], [7], image processing [8], [9], [10], pattern recognition and system identification [11]. The existing subspace clustering algorithms can be divided into four main categories: iterative [12], [13], [14], [15], [16], [17], algebraic [18], [19], [20], [21] statistical [22], [23], and spectral clustering-based methods [24], [25], [26], [27]. The recently developed state-of-the-art subspace clustering algorithms are all spectral clustering-based methods.

*1.1 Related work in spectral clustering-based methods*

There are two main steps involved in the spectral clustering-based methods. The first step is to build a similarity matrix between the data samples and the second step is to use spectral clustering [28], [29], [30] to this similarity matrix to obtain the clustering of the data. Different ways of constructing the similarity matrix lead to different spectral-clustering based algorithms. The spectral clustering-based approaches can be divided into two kinds based on the information used when constructing the similarity matrix which are local spectral clustering-based approaches and global spectral clustering-based approaches. Typical local spectral clustering-based approaches include local subspace affinity (LSA) [24], spectral local best-fit flats (SLBF) [25], [26] and locally linear manifold clustering (LLMC) [27]. These approaches all use local information around each data sample to build a similarity between pairs of samples. These approaches are based on a common assumption: the nearest neighbors of a data sample are from the same subspace as that data sample. These nearest neighbors together with the data sample itself will be used to estimate the local linear subspace of that sample. This assumption certainly has many limitations. The nearby samples are not necessarily from the same subspace while far away samples may be from the same subspace. In addition, they are sensitive to the right choice of the neighborhood size to compute the local information around each sample.

Global spectral clustering-based approaches, on the other hand, try to overcome these difficulties by building more reasonable similarities between data samples using global information. One of the early global spectral clustering-based approaches is spectral curvature clustering (SCC) [31]. SCC uses the concept of polar curvature to define multiway affinities between data samples within an affine subspace. There are many disadvantages of SCC. First, SCC requires knowing the number and dimensions of subspaces and that the dimension of subspaces to be equal. Second, the complexity of building the multiway affinities between data samples grows exponentially with the dimensions of the subspace. The recently proposed global spectral-clustering based approaches use the concept from sparse representation [32], [33], [34] and low-rank approximation [35], [36], [37]. Two most representative approaches are sparse subspace clustering (SSC) [38] and low-rank representation (LRR) [39]. These two algorithms aim to find a representation matrix of the data samples using a dictionary composed of the data itself. In the procedure of the algorithm, a convex optimization problem is formed through which the representation matrix can be solved. Many improvements of SSC and LRR are proposed. There is a main disadvantage of LRR. The dictionary used to represent the data samples is the data itself. However, in practice, data are often contaminated with noises and outliers. Using original data points with noises and outliers as the dictionary is not reasonable. In order to overcome this difficulty, low-rank subspace clustering method (LRSC) [40] is proposed. LRSC uses a clean dictionary to represent the data samples. Using a clean dictionary is more reasonable than using original data samples with noises and outliers and thus higher clustering accuracy can be obtained by LRSC.

*1.2 Incorporating manifold information in data representation and clustering*

|  | Method | Model | Constraints | Subspace Model |
|---|---|---|---|---|
| 1 | PCA [2] | $\min_{V,W} \|X - VW\|_F^2$ | $V^TV = I, (c < p)$ | Single linear subspace |
| 2 | RPCA [3] | $\min_{U,S} \|U\|_* + \lambda \|S\|_1$ | $X = U + S$ | Single linear subspace |
| 3 | NMF [4] | $\min_{V,W} \|X - VW\|_F^2$ | $V \geq 0, W \geq 0, (c < p)$ | Positive single linear subspace |
| 4 | SSC [38] | $\min_{C,E,Z} \|C\|_1 + \lambda_e \|E\|_1 + \frac{\lambda_z}{2}\|Z\|_F^2$ | $X = XC + E + Z, \text{diag}(C) = 0$ | Union of subspaces |
| 5 | LRR [39] | $\min_{Z,E} \|Z\|_* + \lambda \|E\|_{2,1}$ | $X = XZ + E$ | Union of subspaces |
| 6 | LRSC [40] | $\min_{A,E,C} \|C\|_* + \frac{\tau}{2}\|A - AC\|_F^2 + \beta \|E\|_1$ | $X = A + E, C = C^T$ | Union of subspaces |

Linear subspace models **+** Manifold information in graph Laplacian $\mathcal{L}$

Graph Laplacian regularized linear subspace models

|  | Method | Model | Constraints | Parameters | Factors? | Convex? |
|---|---|---|---|---|---|---|
| 7 | GLPCA [46] | $\min_{V,W} \|X - VW\|_F^2 + \gamma \text{tr}(W\mathcal{L}W^T)$ | $V^TV = I, (c < p)$ | $\gamma$ | + | - |
| 8 | RPCAG [47] | $\min_{U,S} \|U\|_* + \lambda \|S\|_1 + \gamma \text{tr}(U\mathcal{L}U^T)$ | $X = U + S$ | $\lambda, \gamma$ | - | + |
| 9 | MMF [48] | $\min_{V,W} \|X - VW\|_F^2 + \gamma \text{tr}(W\mathcal{L}W^T)$ | $V \geq 0, W \geq 0, (c < p)$ | $\gamma$ | + | - |
| 10 | GLSSC [49] | $\min_{C,E,Z} \|C\|_1 + \lambda_e \|E\|_1 + \frac{\lambda_z}{2}\|Z\|_F^2 + \gamma \text{tr}(C\mathcal{L}C^T)$ | $X = XC + E + Z, \text{diag}(C) = 0$ | $\lambda_e, \lambda_z, \gamma$ | - | + |
| 11 | GLLRR [50] | $\min_{Z,E} \|Z\|_* + \lambda \|E\|_{2,1} + \gamma \text{tr}(Z\mathcal{L}Z^T)$ | $X = XZ + E$ | $\lambda, \gamma$ | - | + |
| 12 | **PROPOSED** | $\min_{A,E,C} \|C\|_* + \frac{\tau}{2}\|A - AC\|_F^2 + \beta \|E\|_1 + \gamma \text{tr}(A\mathcal{L}A^T)$ | $X = A + E, C = C^T$ | $\tau, \beta, \gamma$ | + | - |

Figure 1 A summary of the linear subspace models with and without graph regularization. $X \in \mathbb{R}^{p \times n}$ is the matrix of $n$ $p$-dimensional data vectors, $V \in \mathbb{R}^{p \times c}$ and $W \in \mathbb{R}^{c \times n}$ are the learned factors. $U \in \mathbb{R}^{p \times n}$, $A \in \mathbb{R}^{p \times n}$ are the low-rank matrix and $S \in \mathbb{R}^{p \times n}$, $E \in \mathbb{R}^{p \times n}$ are the sparse matrix. $C \in \mathbb{R}^{n \times n}$ and $Z \in \mathbb{R}^{n \times n}$ are the representation matrix. $\|\cdot\|_F$, $\|\cdot\|_*$, $\|\cdot\|_1$ and $\|\cdot\|_{2,1}$ denote the Frobenius, nuclear, $l_1$ and $l_{2,1}$ matrix norms respectively. The data manifold $\mathcal{M}$ information can be leveraged in the form of a discrete graph $G$ using graph Laplacian $\mathcal{L} \in \mathbb{R}^{n \times n}$ resulting in various graph Laplacian regularized linear subspace models.

In the modeling and representation of the high dimensional data samples, linear subspace models are often used. Linear subspace model is only an approximation of the data structure. In real world applications, high dimensional data samples actually lie around a nonlinear manifold. In order to more accurately model the high dimensional data samples and to better preserve local geometric structures embedded in a high dimensional space, manifold learning methods are proposed. Representative manifold learning methods include locally linear embedding (LLE) [41], ISOMAP [42], locality preserving projection (LPP) [43], neighborhood preserving embedding (NPE) [44] and Laplacian Eigenmap (LE) [45]. All these algorithms are based on the idea of the so-called local invariance which aims to estimate geometric and topological properties of an unknown manifold from random points lying around it.

Ideas and methods from manifold learning have been incorporated into the single linear subspace models and subspace clustering models. Single linear subspace models like PCA, RPCA and NMF can be improved by using Laplacian regularization and the corresponding improvements are called graph Laplacian regularized PCA (GLPCA) [46], RPCA on graphs (RPCAG) [47] and manifold regularized matrix factorization (MMF) [48] respectively. Subspace clustering models like SSC and LRR are also improved by adding Laplacian regularization and the corresponding models are called graph Laplacian regularized SSC (GLSSC) [49] and graph Laplacian regularized LRR (GLLRR) [50], [51], [52] respectively. Incorporating manifold information using graph Laplacian in linear subspace models can model the nonlinear properties of the high dimensional data samples and preserve local geometric structures embedded in the high dimensional space. The relationship between linear subspace models and their corresponding models with manifold regularization is presented in Fig. 1. Inspired from the ideas of these manifold regularization methods, a graph Laplacian term is added to the LRSC model to regularize the clean dictionary learned from the original data samples. The corresponding model is called graph Laplacian regularized LRSC (GLLRSC) model and we expect that the model can provide more accurate modeling of the data samples compared to original LRSC and GLLRR.

The organization of the paper is as follows: section 2 presents the algorithmic procedure of original LRSC [40]. Graph Laplacian regularized LRSC is given in section 3. Experimental results and analysis are presented in section 4 followed by some concluding remarks in section 5.

## 2. Low rank subspace clustering [40]

Table 1 The 6 problems involved in low rank subspace clustering and their corresponding data types, constraint types, objective functions and solutions

| Problem | Data type | Constraint type | Objective function | Solution |
|---|---|---|---|---|
| $P_1$ | Uncorrupted data | Relaxed constraint | $\min_{C} \|C\|_* + \frac{\tau}{2} \|A - AC\|_F^2$ s.t. $C = C^T$ | Closed form |
| $P_2$ | Uncorrupted data | Exact constraint | $\min_{C} \|C\|_*$ s.t. $A = AC$ and $C = C^T$ | Closed form |
| $P_3$ | Noisy data | Relaxed constraint | $\min_{A,C} \|C\|_* + \frac{\tau}{2} \|A - AC\|_F^2 + \frac{\alpha}{2} \|X - A\|_F^2$ s.t. $C = C^T$ | Closed form, polynomial thresholding |
| $P_4$ | Noisy data | Exact constraint | $\min_{A,C} \|C\|_* + \frac{\alpha}{2} \|X - A\|_F^2$ s.t. $A = AC$ and $C = C^T$ | Closed form, hard thresholding |
| $P_5$ | Corrupted data | Relaxed constraint | $\min_{A,C,E} \|C\|_* + \frac{\tau}{2} \|A - AC\|_F^2 + \frac{\alpha}{2} \|X - A - E\|_F^2 + \beta \|E\|_1$ s.t. $C = C^T$ | Iterative polynomial thresholding or ADMM |
| $P_6$ | Corrupted data | Exact constraint | $\min_{A,C,E} \|C\|_* + \frac{\alpha}{2} \|X - A - E\|_F^2 + \beta \|E\|_1$ s.t. $A = AC$ and $C = C^T$ | Iterative hard thresholding or ADMM |

Low rank subspace clustering actually includes 6 optimization problems in total. These 6 optimization problems are formulated by considering different combinations of data types and constraint types. There are 3 data types which are uncorrupted data, noisy data and corrupted data. There are 2 constraint types which are relaxed constraint and exact constraint. Based on these 3 data types and 2 constraint types, there are total 6 optimization problems. These problems together with their corresponding data types, constraint types, objective functions are listed in Table 1. The solutions to these 6 optimization problems are also listed in the table. In this section we briefly review them and give the solutions to these 6 problems. For more details, please refer to [40].

Problem $P_1$ and problem $P_2$ are similar to each other. Problem $P_1$ and problem $P_2$ both consider the situation where the data are uncorrupted. Uncorrupted data means that the data samples are not corrupted by Gaussian noises or gross errors. The constraint type in problem $P_1$ and problem $P_2$ is not the same. The constraint type in problem $P_1$ is relaxed constraint which means that the data samples are approximately represented by the data samples themselves. The constraint type in problem $P_2$ is exact constraint which means that the data samples are exactly represented by the data samples themselves. The objective functions of problem $P_1$ and problem $P_2$ are given as follows respectively

$(P_1)$ $$\min_{C} \|C\|_* + \frac{\tau}{2}\|A - AC\|_F^2 \quad \text{s.t.} \quad C = C^T$$

$(P_2)$ $$\min_{C} \|C\|_* \quad \text{s.t.} \quad A = AC \text{ and } C = C^T$$

where $A \in \mathbb{R}^{p \times n}$ is the data matrix and $C \in \mathbb{R}^{n \times n}$ is the representation matrix. Notice from the objective function of problem $P_1$ and problem $P_2$ the similarity and difference between them. When $\tau = \infty$, problem $P_1$ becomes problem $P_2$. The representation matrix $C$ in problem $P_1$ and problem $P_2$ can be solved in closed form from the singular value decomposition (SVD) of $A$. Assume the SVD of $A$ is $A = U\Lambda V^T$. The optimal solution to $P_1$ is given by

$$C = V\mathcal{P}_\tau(\Lambda)V^T = V_1\left(I - \frac{1}{\tau}\Lambda_1^{-2}\right)V_1^T \tag{1}$$

where the operator $\mathcal{P}_\tau$ acts on diagonal entries of $\Lambda$ as

$$\mathcal{P}_\tau(x) = \begin{cases} 1 - \frac{1}{\tau x^2} & x > \frac{1}{\sqrt{\tau}} \\ 0 & x \leq \frac{1}{\sqrt{\tau}} \end{cases} \tag{2}$$

and $U = [U_1, U_2]$, $\Lambda = \text{diag}(\Lambda_1, \Lambda_2)$ and $V = [V_1, V_2]$ are partitioned according to the sets $I_1 = \left\{i : \lambda_i > \frac{1}{\sqrt{\tau}}\right\}$ and $I_2 = \left\{i : \lambda_i \leq \frac{1}{\sqrt{\tau}}\right\}$. The singular values of $A$ are placed as the diagonal entries of $\Lambda = \text{diag}(\{\lambda_i\})$ in decreasing order. When $\tau = \infty$, the operator $\mathcal{P}_\tau$ becomes

$$\mathcal{P}_{\tau=\infty}(x) = \begin{cases} 1 & x > 0 \\ 0 & x \leq 0 \end{cases}$$

and the corresponding optimal solution to problem $P_2$ is $C = V_1 V_1^T$ where $V = [V_1, V_2]$ is partitioned according to the sets $I_1 = \{i : \lambda_i > 0\}$ and $I_2 = \{i : \lambda_i = 0\}$.

Just like the relationship between problem $P_1$ and problem $P_2$, problem $P_3$ and problem $P_4$ are similar to each other. Problem $P_3$ and problem $P_4$ both consider the situation where the data are noisy. Noisy data means that the data samples are only corrupted by Gaussian noises but not corrupted by gross errors. The constraint type in problem $P_3$ is relaxed constraint while the constraint type in problem $P_4$ is exact constraint. When $\tau = \infty$, problem $P_3$ becomes problem $P_4$. The objective function of problem $P_3$ and problem $P_4$ are given as follows respectively

$(P_3)$ $$\min_{A, C} \|C\|_* + \frac{\tau}{2}\|A - AC\|_F^2 + \frac{\alpha}{2}\|X - A\|_F^2 \quad \text{s.t.} \quad C = C^T$$

$(P_4)$ $$\min_{A,C} \|C\|_* + \frac{\alpha}{2}\|X - A\|_F^2 \text{ s.t. } A = AC \text{ and } C = C^T$$

where $X \in \mathbb{R}^{p \times n}$ is the data matrix, $A \in \mathbb{R}^{p \times n}$ is the learned clean dictionary and $C \in \mathbb{R}^{n \times n}$ is the representation matrix. Since the existence of the product $AC$, the objective function in $P_3$ is not convex in $(A, C)$. The optimal solutions for both $A$ and $C$ in problem $P_3$ and problem $P_4$ can be computed in closed form from the SVD of $X$ and the optimal solutions are unique. Assume the SVD of $X$ is $X = U\Sigma V^T$. The optimal solution to problem $P_3$ is given by

$$A = U\Lambda V^T \text{ and } C = V\mathcal{P}_\tau(\Lambda)V^T \tag{3}$$

where each entry of $\Lambda = diag(\{\lambda_1,...,\lambda_n\})$ is obtained from each entry of $\Sigma = diag(\{\sigma_1,...,\sigma_n\})$ as the solutions to

$$\sigma = \psi(\lambda) = \begin{cases} \lambda + \frac{1}{\alpha\tau}\lambda^{-3} & \text{if } \lambda > \frac{1}{\sqrt{\tau}} \\ \lambda + \frac{\tau}{\alpha}\lambda & \text{if } \lambda \leq \frac{1}{\sqrt{\tau}} \end{cases} \tag{4}$$

that minimize

$$\phi(\lambda, \sigma) = \frac{\alpha}{2}(\sigma - \lambda)^2 + \begin{cases} 1 - \frac{1}{2\tau}\lambda^{-2} & \text{if } \lambda > \frac{1}{\sqrt{\tau}} \\ \frac{\tau}{2}\lambda^2 & \text{if } \lambda \leq \frac{1}{\sqrt{\tau}} \end{cases} \tag{5}$$

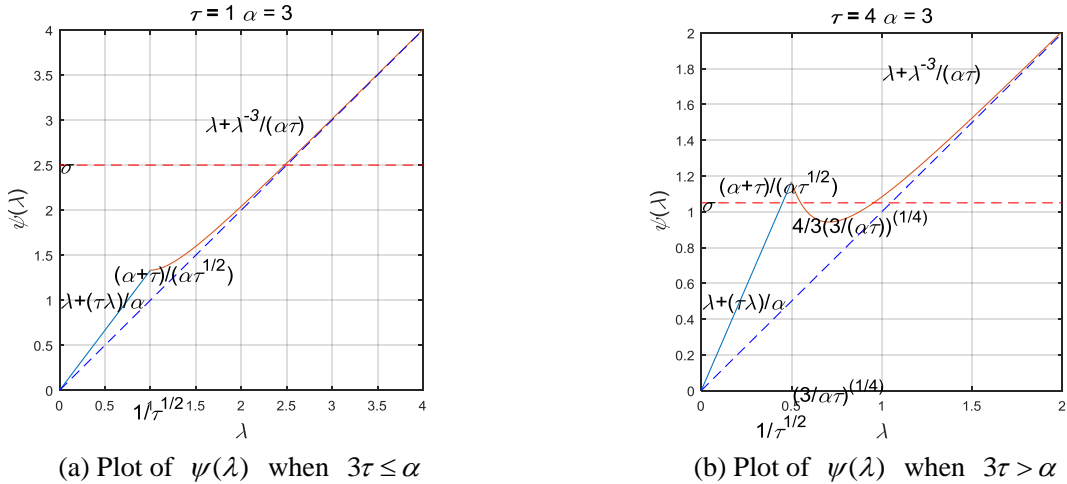

(a) Plot of $\psi(\lambda)$ when $3\tau \leq \alpha$  (b) Plot of $\psi(\lambda)$ when $3\tau > \alpha$

Figure 2 Plots of $\psi(\lambda)$

The plots of $\psi(\lambda)$ are shown in Fig. 2. Notice the difference between Fig. 2(a) and Fig. 2(b). When $3\tau \leq \alpha$, function $\psi(\lambda)$ is a strictly increasing function. Thus the solution for $\lambda$ is unique (see Fig. 2(a)). When $3\tau > \alpha$, function $\psi(\lambda)$ first increases in the interval $\left[0, \frac{1}{\sqrt{\tau}}\right]$, then decreases in the interval $\left[\frac{1}{\sqrt{\tau}}, \sqrt[4]{\frac{3}{\alpha\tau}}\right]$ and then again increases in the interval $\left[\sqrt[4]{\frac{3}{\alpha\tau}}, \infty\right]$. There could be three possible solutions in this situation (see Fig. 2(b)). In general, only one of these solutions is the global minimizer. This solution can be obtained by applying a polynomial thresholding operator $\lambda = \mathcal{P}_{\alpha,\tau}(\sigma)$ to the singular values of $X$. The polynomial thresholding operator

can be defined as

$$\lambda = \mathcal{P}_{\alpha,\tau}(\sigma) = \begin{cases} \beta_1(\sigma) = \sigma & \text{if } \sigma > \sigma_* \\ \beta_3(\sigma) = \dfrac{\alpha}{\alpha+\tau}\sigma & \text{if } \sigma \leq \sigma_* \end{cases} \quad (6)$$

where the parameter $\sigma_*$ is given as follows

$$\sigma_* = \sqrt{\dfrac{\alpha+\tau}{\alpha\tau}} + \sqrt{\dfrac{\alpha+\tau}{\alpha^2\tau}} \quad (7)$$

When $\tau = \infty$, the polynomial thresholding operator becomes

$$\lambda = \mathcal{P}_{\alpha,\tau=\infty}(\sigma) = \begin{cases} \beta_1(\sigma) = \sigma & \text{if } \sigma > \sigma_* \\ \beta_3(\sigma) = 0 & \text{if } \sigma \leq \sigma_* \end{cases}$$

where $\sigma_* = \sqrt{\dfrac{2}{\alpha}}$. In this case, the polynomial thresholding operator becomes a hard thresholding operator. Thus the optimal solution to problem $P_4$ is given by

$$A = U_1 \Sigma_1 V_1^T \text{ and } C = V_1 V_1^T \quad (8)$$

where $\Sigma_1$ contains the singular values of $X$ that are larger than $\sqrt{\dfrac{2}{\alpha}}$ and $U_1$ and $V_1$ contain the corresponding singular vectors.

Just like the relationship between problem $P_1$ and problem $P_2$ and the relationship between problem $P_3$ and problem $P_4$, problem $P_5$ and problem $P_6$ are similar to each other. Problem $P_5$ and problem $P_6$ both consider the situation where the data are corrupted. Corrupted data means that the data samples are simultaneously corrupted by Gaussian noises and gross errors. The constraint type in problem $P_5$ is relaxed constraint while the constraint type in problem $P_6$ is exact constraint. When $\tau = \infty$, problem $P_5$ becomes problem $P_6$. The objective functions of problem $P_5$ and problem $P_6$ are given as follows respectively

$(P_5)$ $\quad \min\limits_{A,C,E} \|C\|_* + \dfrac{\tau}{2}\|A - AC\|_F^2 + \dfrac{\alpha}{2}\|X - A - E\|_F^2 + \beta\|E\|_1 \text{ s.t. } C = C^T$

$(P_6)$ $\quad \min\limits_{A,C,E} \|C\|_* + \dfrac{\alpha}{2}\|X - A - E\|_F^2 + \beta\|E\|_1 \text{ s.t. } A = AC \text{ and } C = C^T$

Problem $P_5$ can be solved using iterative polynomial thresholding (IPT) while problem $P_6$ can be solved using iterative hard thresholding. The optimization algorithm of problem $P_5$ mainly contains two steps. The first step is applying polynomial thresholding operator to $X - E_{(k)}$ to obtain $A_{(k+1)}$. The second step is applying shrinkage-thresholding operator to $X - A_{(k+1)}$ to obtain $E_{(k+1)}$. The optimization algorithm alternates between the above two steps. The initial condition can be set as $A_{(0)} = X$ and $E_{(0)} = 0$. The procedure of the algorithm can be given as follows.

**Algorithm** IPT for solving problem $P_5$ ( $X = A + E + G$ )

**Input:** data matrix $X$, parameter $\tau$, $\alpha$, $\beta$

**Initialize:** $A_{(0)} = X$, $E_{(0)} = 0$, $\varepsilon_1 = 10^{-8}$ and $k=0$

**while** not converged **do**

    **1.** applying polynomial thresholding operator to $X - E_{(k)}$ to obtain $A_{(k+1)}$

$$(U_{(k)}, \Sigma_{(k)}, V_{(k)}) = \text{svd}(X - E_{(k)})$$

$$A_{(k+1)} = U_{(k)} \mathcal{P}_{\alpha, \tau}(\Sigma_{(k)}) V_{(k)}^T$$

    **2.** applying shrinkage-thresholding operator to $X - A_{(k+1)}$ to obtain $E_{(k+1)}$

$$E_{(k+1)} = \mathcal{S}_{\beta\alpha^{-1}}(X - A_{(k+1)})$$

    **3.** check the convergence conditions

$$\| A_{(k+1)} - A_{(k)} \|_\infty < \varepsilon_1 \text{ and } \| E_{(k+1)} - E_{(k)} \|_\infty < \varepsilon_1$$

**end while**

$$(U_{(k+1)}, \Sigma_{(k+1)}, V_{(k+1)}) = \text{svd}(X - E_{(k+1)})$$

$$C = V_{(k+1)} \mathcal{P}_\tau(\mathcal{P}_{\alpha, \tau}(\Sigma_{(k+1)})) V_{(k+1)}^T$$

**Output:** Matrix $C$, $A = A_{(k+1)}$ and $E = E_{(k+1)}$

---

Notice that the representation matrix $C$ can be obtained from $A$ upon convergence. In the original problem $P_5$, the data matrix can be decomposed as $X = A + E + G$, where $E$ represents the gross errors and $G$ represents the Gaussian noises. When the data samples are only corrupted by gross errors and not corrupted by Gaussian noises, the data matrix can be written as $X = A + E$. In this case, problem $P_5$ can be solved by the augmented Lagrange multipliers method. The objective function of problem $P_5$ can be formulated as

$$\min_{A, C, E} \| C \|_* + \frac{\tau}{2} \| A - AC \|_F^2 + \frac{\mu}{2} \| X - A - E \|_F^2 + <Y, X - A - E> + \beta \| E \|_1 \text{ s.t. } C = C^T \quad (9)$$

where $Y$ is the Lagrange multiplier. This problem can be solved by alternating direction method of multipliers (ADMM). The procedure of the algorithm can be given as follows.

**Algorithm** ADMM for solving problem $P_5$ ( $X = A + E$ )

**Input:** data matrix $X$, parameter $\tau$, $\beta$

**Initialize:** $A_{(0)} = X$, $E_{(0)} = 0$, $Y_{(0)} = 0$, $\varepsilon_1 = 10^{-8}$, $\mu_{(0)} = 10^{-6}$, $\rho = 1.1$ and $k=0$

**while** not converged **do**

    **1.** fix the others and update $A$ by

$$(U_{(k)}, \Sigma_{(k)}, V_{(k)}) = \text{svd}(X - E_{(k)} + \mu_{(k)}^{-1} Y_{(k)})$$

$$A_{(k+1)} = U_{(k)} \mathcal{P}_{\mu_{(k)}, \tau}(\Sigma_{(k)}) V_{(k)}^T$$

    **2.** fix the others and update $E$ by

$$E_{(k+1)} = \mathcal{S}_{\beta \mu_{(k)}^{-1}}(X - A_{(k+1)} + \mu_{(k)}^{-1} Y_{(k)})$$

    **3.** update the multipliers

$$Y_{(k+1)} = Y_{(k)} + \mu_{(k)}(X - A_{(k+1)} - E_{(k+1)})$$

    **4.** update the parameter by $\mu_{(k+1)} = \rho \mu_{(k)}$

    **5.** check the convergence conditions

$$\| X - A_{(k+1)} - E_{(k+1)} \|_\infty < \varepsilon_1$$

**end while**

$$(U_{(k+1)}, \Sigma_{(k+1)}, V_{(k+1)}) = \text{svd}(X - E_{(k+1)} + \mu_{(k+1)}^{-1} Y_{(k+1)})$$

$$C = V_{(k+1)} \mathcal{P}_\tau(\mathcal{P}_{\alpha, \tau}(\Sigma_{(k+1)})) V_{(k+1)}^T$$

**Output:** Matrix $C$, $A = A_{(k+1)}$ and $E = E_{(k+1)}$

Notice that the representation matrix $C$ is also obtained from $A$ upon convergence. The optimization algorithm of problem $P_6$ is similar to that of problem $P_5$ and the only difference is that the polynomial thresholding operator $\mathcal{P}_{\alpha, \tau}$ is replaced by the hard thresholding operator.

## 3. Graph Laplacian regularized low rank subspace clustering

In order to incorporate the data manifold information in the form of a discrete graph into the LRSC framework, a graph Laplacian regularization term is added to the objective function. The graph smoothness constraint can be applied to different terms in the original objective function. In GLPCA, MMF, GLSSC and GLLRR, the graph smoothness constraint is applied to the representation coefficients. In RPCAG, the graph smoothness constraint is applied to the low rank matrix learned from the data matrix. In the model proposed in this paper, the graph smoothness constraint is applied to the clean dictionary learn from the data matrix and the way incorporating graph Laplacian regularization term in the model proposed in this paper is similar to RPCAG. The graph Laplacian regularized LRSC with corrupted data and relaxed constraint is given as follows

(GL-$P_5$) $\quad \min_{A, C, E} \| C \|_* + \frac{\tau}{2} \| A - AC \|_F^2 + \beta \| E \|_1 + \gamma \,\text{tr}(A \mathcal{L} A^T)$ s.t. $X = A + E$ and $C = C^T$

where $\mathcal{L}$ is the unnormalized graph Laplacian matrix and $\mathcal{L}$ can be computed by

$$\mathcal{L} = D - W \tag{10}$$

where $W$ is a similarity matrix and $W_{ij}$ represents the similarity between data points $x_i$ and $x_j$, $D$ is a diagonal matrix and $D_{ii} = \sum_j W_{ij}$ which measures the similarity of the data point $x_i$ with all the other data points in the data matrix. The similarity matrix can be defined in different ways. Two most widely used definition of the similarity matrix are the $K$-nearest neighbor similarity matrix and $\varepsilon$-neighborhood similarity matrix. The similarity matrix adopted in this paper is the $K$-nearest neighbor similarity matrix and its definition is given as follows

$$W_{ij} = \begin{cases} 1 & \text{if } x_j \text{ is among the } K \text{ nearest neighbors of } x_i \\ 0 & \text{otherwise} \end{cases} \tag{11}$$

Note that the $K$-nearest neighbor similarity matrix is not symmetric. The problem GL-$P_5$ can be solved by using ADMM. First the objective function of problem GL-$P_5$ can be written as

$$\min_{J,A,C,E} \|C\|_* + \frac{\tau}{2} \|A - AC\|_F^2 + \beta \|E\|_1 + \gamma \operatorname{tr}(J\mathcal{L}J^T) \text{ s.t. } X = A + E, J = A \text{ and } C = C^T \tag{12}$$

Then the augmented Lagrange multipliers are added to the above objective function making it an unconstraint optimization problem

$$\min_{J,A,C,E} \|C\|_* + \frac{\tau}{2} \|A - AC\|_F^2 + \beta \|E\|_1 + \gamma \operatorname{tr}(J\mathcal{L}J^T) + <Y_1, X - A - E> + \frac{\mu_1}{2} \|X - A - E\|_F^2 + <Y_2, J - A> + \frac{\mu_2}{2} \|J - A\|_F^2$$
(13)

where $Y_1$ and $Y_2$ are the Lagrange multipliers. The above objective function can be solved by fixing the other variables and optimize it with only one variable. Fixing variables $A$, $C$, $E$ and optimize it with variable $J$

$$J_{(k+1)} = \arg\min_J \gamma \operatorname{tr}(J\mathcal{L}J^T) + <Y_{2,(k)}, J - A_{(k)}> + \frac{\mu_{2,(k)}}{2} \|J - A_{(k)}\|_F^2 \tag{14}$$

$J_{(k+1)}$ can be solved in closed form

$$J_{(k+1)} = (\mu_2 A_{(k)} - Y_{2,(k)})(2\gamma\mathcal{L} + \mu_2 I)^{-1} \tag{15}$$

Fixing $J$, $C$, $E$ and optimize it with variable $A$

$$A_{(k+1)} = \arg\min_A \frac{\tau}{2} \|A - AC_{(k)}\|_F^2 + <Y_{1,(k)}, X - A - E_{(k)}> + \frac{\mu_{1,(k)}}{2} \|X - A - E_{(k)}\|_F^2 + <Y_{2,(k)}, J_{(k+1)} - A> + \frac{\mu_{2,(k)}}{2} \|J_{(k+1)} - A\|_F^2$$
(16)

**Algorithm 1** Optimization algorithm for solving GL-$P_5$

**Input:** data matrix $X$, parameter $\tau$, $\beta$, $\gamma$

**Initialize:** $J_{(0)} = 0$, $A_{(0)} = 0$, $E_{(0)} = 0$, $Y_{1,(0)} = 0$, $Y_{2,(0)} = 0$, $\mu_{1,(0)} = 10^{-6}$, $\mu_{2,(0)} = 10^{-6}$, $\mu_{max} = 10^{6}$, $\rho = 1.1$, $\varepsilon_1 = 10^{-8}$ and $k=0$

**while** not converged **do**

    **1.** fix the others and update $J$ by

$$J_{(k+1)} = (\mu_{2,(k)}A_{(k)} - Y_{2,(k)})(2\gamma \mathcal{L} + \mu_{2,(k)}I)^{-1}$$

    **2.** fix the others and update $A$ by

$$(U_{(k)}, \Sigma_{(k)}, V_{(k)}) = \text{svd}\left( \frac{\mu_{1,(k)}\left(X - E_{(k)} + \frac{Y_{1,(k)}}{\mu_{1,(k)}}\right) + \mu_{2,(k)}\left(J_{(k+1)} + \frac{Y_{2,(k)}}{\mu_{2,(k)}}\right)}{\mu_{1,(k)} + \mu_{2,(k)}} \right)$$

$$A_{(k+1)} = U_{(k)} \mathcal{P}_{\mu_{1,(k)}+\mu_{2,(k)},\tau}(\Sigma_{(k)}) V_{(k)}^T$$

    **3.** fix the others and update $E$ by

$$E_{(k+1)} = \mathcal{S}_{\beta\mu_{1,(k)}^{-1}}(X - A_{(k+1)} + \mu_{1,(k)}^{-1}Y_{1,(k)})$$

    **4.** update the multipliers

$$Y_{1,(k+1)} = Y_{1,(k)} + \mu_{1,(k)}(X - A_{(k+1)} - E_{(k+1)})$$

$$Y_{2,(k+1)} = Y_{2,(k)} + \mu_{2,(k)}(J_{(k+1)} - A_{(k+1)})$$

    **5.** update the parameter $\mu_1$ and $\mu_2$ by $\mu_{1,(k+1)} = \min(\rho\mu_{1,(k)}, \mu_{max})$ and $\mu_{2,(k+1)} = \min(\rho\mu_{2,(k)}, \mu_{max})$

    **6.** check the convergence conditions

$$\|X - A_{(k+1)} - E_{(k+1)}\|_\infty < \varepsilon_1 \text{ and } \|J_{(k+1)} - A_{(k+1)}\|_\infty < \varepsilon_1$$

    **7.** $k=k+1$

**end while**

$$(U_{(k+1)}, \Sigma_{(k+1)}, V_{(k+1)}) = \text{svd}\left( \frac{\mu_{1,(k+1)}\left(X - E_{(k+1)} + \frac{Y_{1,(k+1)}}{\mu_{1,(k+1)}}\right) + \mu_{2,(k+1)}\left(J_{(k+1)} + \frac{Y_{2,(k+1)}}{\mu_{2,(k+1)}}\right)}{\mu_{1,(k+1)} + \mu_{2,(k+1)}} \right)$$

$$C = V_{(k+1)} \mathcal{P}_\tau(\mathcal{P}_{\mu_{1,(k+1)}+\mu_{2,(k+1)},\tau}(\Sigma_{(k+1)})) V_{(k+1)}^T$$

**Output:** Matrix $C$, $A = A_{(k+1)}$ and $E = E_{(k+1)}$

$$A_{(k+1)} = \arg\min_A \frac{\tau}{2}\|A - AC_{(k)}\|_F^2 + \frac{\mu_{1,(k)} + \mu_{2,(k)}}{2}\left\| A - \frac{\mu_1\left(X - E_{(k)} + \frac{Y_{1,(k)}}{\mu_{1,(k)}}\right) + \mu_2\left(J_{(k+1)} + \frac{Y_{2,(k)}}{\mu_{2,(k)}}\right)}{\mu_{1,(k)} + \mu_{2,(k)}} \right\|_F^2 \quad (17)$$

Optimization problem (17) is equivalent to problem $P_3$ and it can be solved by applying polynomial thresholding operation.

$$(U_{(k)}, \Sigma_{(k)}, V_{(k)}) = \text{svd}\left(\frac{\mu_1\left(X - E_{(k)} + \frac{Y_{1,(k)}}{\mu_1}\right) + \mu_2\left(J_{(k+1)} + \frac{Y_{2,(k)}}{\mu_2}\right)}{\mu_1 + \mu_2}\right)$$

$$A_{(k+1)} = U_{(k)} \mathcal{P}_{\mu_1+\mu_2,\tau}(\Sigma_{(k)}) V_{(k)}^T \tag{18}$$

Fixing $J, C, A$ and optimize it with variable $E$

$$E_{(k+1)} = \arg\min_{E} \beta \|E\|_1 + <Y_{1,(k)}, X - A_{(k+1)} - E> + \frac{\mu_{1,(k)}}{2}\|X - A_{(k+1)} - E\|_F^2 \tag{19}$$

$$E_{(k+1)} = \mathcal{S}_{\beta\mu_{1,(k)}^{-1}}(X - A_{(k+1)} + \mu_{1,(k)}^{-1} Y_{1,(k)}) \tag{20}$$

As before, since the optimization procedure doesn't include the matrix $C$, matrix $C$ can be obtained upon convergence of the algorithm. The whole algorithm for optimizing problem GL-$P_5$ is given in algorithm 1.

The graph Laplacian regularization term can also be added to the objective function of problem $P_6$. The resulting objective function is given as follows

(GL-$P_6$) $\quad\quad \min_{A,C,E} \|C\|_* + \beta\|E\|_1 + \gamma \text{tr}(A\mathcal{L}A^T)$ s.t. $X = A + E, A = AC$ and $C = C^T$

The solution of the above objective function is similar to algorithm 1 where the only difference is the replacement of polynomial thresholding operation in step 2 by hard thresholding operation.

## 4. Experimental results and analysis

In this section, we evaluate the performance of the proposed GLLRSC ($P_5$) and GLLRSC ($P_6$) in dealing with two real-world problems which are clustering images of human faces and clustering images of hand written digits. The performance of the proposed method is compared with the best state-of-the-art subspace clustering algorithms: LSA [24], SLBF [25], LLMC [27], SCC [31], MSL [16], SSC [38], LRR [39] and LRSC ($P_5$ and $P_6$) [40]. Here LRSC ($P_5$) and LRSC ($P_6$) are solved by the ADMM method. LSA, SLBF, LLMC are local spectral clustering-based algorithms and SCC, SSC, LRR and LRSC are global spectral clustering-based algorithms. MSL can be considered as a kind of iterative statistical method.

**Implementation details.** In GLLRSC ($P_5$), for face clustering, the parameters are chosen as follows: $\tau = 0.2$, $\beta = 10^{-6}$, $\gamma = 10^{-3}$. 10-neareast neighbors are chosen to compute the similarity matrix. For handwritten digit clustering, the parameters are chosen as follows: $\tau = 0.2$, $\beta = 10^{-6}$, $\gamma = 10^{-2}$. 5-nearest neighbors are chosen to compute the similarity matrix. In GLLRSC ($P_6$), for face clustering, the parameters are chosen as follows: $\beta = 10^{-6}$, $\gamma = 10^{-3}$. 10-nearest neighbors are chosen to compute the similarity matrix. For handwritten digit clustering, the parameters are chosen as follows: $\beta = 10^{-6}$, $\gamma = 10^{-2}$. 5-nearest neighbors are chosen to compute the similarity matrix. We use the codes provided by their authors for the comparison algorithms. For LSA, $K$=15 nearest neighbors and subspace dimension $d$=9 are chosen to fit local subspaces for face clustering and $K$=5 nearest neighbors and subspace dimension $d$=9 are chosen for handwritten digit clustering. For SLBF, subspace dimension $d$=9 is chosen to fit local subspaces for face clustering and handwritten digit clustering. The start size and step size of local neighborhood selection is respectively set as 10 and 5 for face clustering and 2 and 2 for handwritten digit clustering. For LLMC, $K$=10 nearest neighbors are chosen for face clustering and $K$=5 nearest neighbors are chosen for handwritten digit clustering. For SCC, subspace dimension $d=9$ is chosen to fit local subspaces and the affine SCC clustering version is used for face clustering and handwritten digit clustering. For MSL, the initial segmentation is provided by the shape interaction matrix method and subspace dimension $d=9$ is chosen to fit local subspaces for face clustering and handwritten digit clustering. For LRR, the suggested weighting parameter $\lambda = 1/\sqrt{\log(n)}$ is used [53]. For SSC, the optimization program with affine constraint and noise term and sparse outlying term is used with the weighting parameter $\lambda_z = 800/\mu_z$ for the noise term and $\lambda_e = 20/\mu_e$ for the sparse outlying term as suggested by the authors. For LRSC ($P_5$), the parameters are set as follows: $\tau = 0.2$, $\beta = 10^{-6}$. For LRSC ($P_6$), the parameter is set as follows: $\beta = 10^{-6}$. Finally, as LSA, SLBF, LLMC, SCC and MSL need to know the number of subspaces a prior and the estimation of the number of subspaces from the eigenspectrum of the graph Laplacian in the noisy setting is often unreliable [38], to have a fair comparison we provide the number of subspaces as an input to all the algorithms.

**Datasets.** For the face clustering problem, we consider the Extended Yale B dataset [54], which consists of face images of 38 human subjects, where images of each subject lie in a low-dimensional manifold. For the handwritten digit clustering problem, we consider the MNIST dataset and USPS dataset. The MNIST dataset of handwritten digits has a training set of 60000 examples and a test set of 10000 examples. The digits have been size-normalized and centered in a fixed-size image. The USPS dataset consists of a training set with 7291 images and a test set with 2007 images. The digits have also been size-normalized. The images of each digit also lie in a low-dimensional manifold. The details of all datasets used in the experiment are provided in Table 2.

Table 2 Details of the datasets used for clustering experiments in this work

| Dataset | Training samples | Test samples | Dimension | Classes |
|---|---|---|---|---|
| Extended Yale B | / | 2432 | $192 \times 168$ | 38 |
| MNIST | 60000 | 10000 | $28 \times 28$ | 10 |
| USPS | 7291 | 2007 | $16 \times 16$ | 10 |

*4.1 Face clustering*

The face images in the Extended Yale B dataset are acquired under a fixed pose with varying lighting conditions. It has been shown that, under the Lambertian assumption, images of a subject with a fixed pose and vary lighting conditions lie close to a linear subspace of dimension 9 [38]. Thus, the collection of face images of multiple subjects lies close to a union of 9D subspaces [38], [39].

In this section, the clustering performance of GLLRSC ($P_5$) and GLLRSC ($P_6$) as well as the state-of-the-art methods on the Extended Yale B dataset [54] are evaluated and compared. The dataset consists of $192 \times 168$ pixel cropped face images of $n=38$ individuals, where there are $N_i = 64$ frontal face images for each subject acquired under various lighting conditions. As done in the literature [38], we also downsample the images to $48 \times 42$ pixels to reduce the computational cost and the memory requirements of all algorithms. Thus each data point is a 2016D vector, hence $p=2016$.

We divide the 38 subjects into four groups as done in [38], where the first three groups correspond to subjects 1 to 10, 11 to 20, 21 to 30, and the fourth group corresponds to subjects 31 to 38. The four groups are referred to as group 1-4 respectively. The authors in [38] consider all choices of $n \in \{2,3,5,8,10\}$ subjects for each of the first three groups and $n \in \{2,3,5,8\}$ for the last group. In this paper, we only consider $n=10$ for each of the first three groups and $n=8$ for the last group since this is the most challenging situation for face clustering. In [38], the authors reported clustering results in three different settings: the first is applying RPCA separately on each subject and then use the clustering algorithm, the second is applying PRCA simultaneously on all subjects and then use the clustering algorithm and the third is directly using the clustering algorithm to the original data points. In this paper, the first and second settings are not considered and all the clustering algorithms are directly applied to the original data points. The clustering accuracy results of different algorithms are presented in Table 3.

Table 3 Clustering accuracy (%) of different algorithms on the Extended Yale B dataset without preprocessing the data

| Algorithm | Group 1 | Group 2 | Group 3 | Group 4 |
| --- | --- | --- | --- | --- |
| LSA [24] | 25.00 | 25.68 | 25.31 | 30.44 |
| SLBF [25] | 37.26 | 38.36 | 34.06 | 40.12 |
| LLMC [27] | 40.09 | 43.18 | 32.34 | 38.54 |
| SCC [31] | 29.25 | 20.39 | 30.31 | 30.63 |
| MSL [16] | 64.94 | 59.55 | 71.41 | 42.69 |
| SSC [38] | 83.18 | 89.89 | 94.22 | 78.06 |
| LRR [39] | 95.44 | 85.23 | 96.72 | 90.32 |
| LRSC ($P_5$) [40] | 75.00 | 82.18 | 95.63 | 87.95 |
| LRSC ($P_6$) [40] | 91.67 | 82.18 | 94.69 | 89.53 |
| GLLRSC ($P_5$) | 84.12 | 87.16 | 97.19 | 92.89 |
| GLLRSC ($P_6$) | 84.28 | 88.76 | 97.34 | 92.69 |

From the results we make the following conclusions. Local spectral clustering-based algorithms such as LSA, SLBF and LLMC obtain a relatively low clustering accuracy on the dataset. The local linear subspace of a data sample is estimated using the K nearest neighbors of that data sample in local spectral clustering-based algorithms. As for the Extended Yale B dataset, there are a relatively large number of points near the intersection of subspaces; the local neighborhood of a sample must contain a lot of samples from other subspaces. Thus the local subspace estimated cannot reflect the true local subspace and thus the similarity built based on this local subspace cannot reflect the true similarity between data samples.

Global spectral clustering-based algorithm such as SCC also obtains low clustering accuracy. Global spectral clustering-based algorithm doesn't rely on the nearest neighbors. However, the noises and outliers exist in the dataset will affect the estimation of the polar curvature and thus the similarity built based on this polar curvature cannot reflect the true similarity between the data samples.

Iterative statistical method such as MSL obtains relative low clustering accuracy. This method needs an initial clustering of the dataset. This method then uses an iterate method to refine the initial clustering results. This method can easily get stuck in local minimum around the initial segmentation and the clustering accuracy is thus highly dependent on the initial clustering.

Global spectral clustering-based algorithms such as LRR and SSC obtain more accurate clustering results while LRR gets higher clustering accuracy than SSC. LRR and SSC both exploit the global structure of the dataset and handle noises and outliers explicitly. The representation matrix obtained can reflect the relationship of the data sample with other samples and the similarity matrix is more accurate.

LRSC ($P_5$) and LRSC ($P_6$) obtain similar clustering results as that of LRR and the clustering accuracy of LRSC ($P_6$) is a little higher than that of LRSC ($P_5$). LRSC uses the clean dictionary to represent the data samples. Comparing the clustering accuracy of LRR and LRSC, we find that the clustering accuracy of LRSC is a bit lower than that of LRR. The clean dictionary learned may still contain some noises and outliers. GLLRSC ($P_5$) and GLLRSC ($P_6$) obtain the highest clustering accuracy among all the algorithms and the clustering accuracy of GLLRSC ($P_6$) is also a little higher than that of GLLRSC ($P_5$). Incorporating manifold information into the objective function of LRSC can improve the clustering accuracy of LRSC and the learned dictionary from the data matrix is more close to a clean dictionary than the original LRSC algorithm. Using hard thresholding is also better than using polynomial thresholding in this dataset as comparing the clustering results in the table.

*4.2 Handwritten digit clustering*

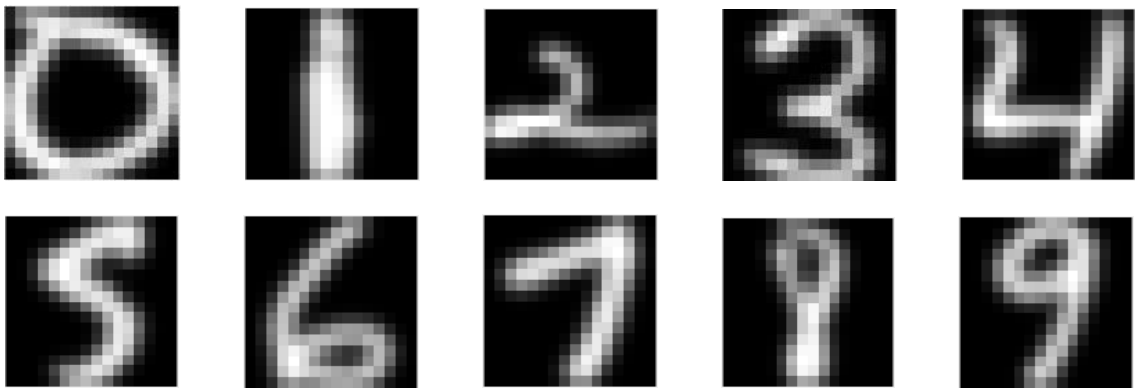

Figure 3 Some data samples from the USPS dataset

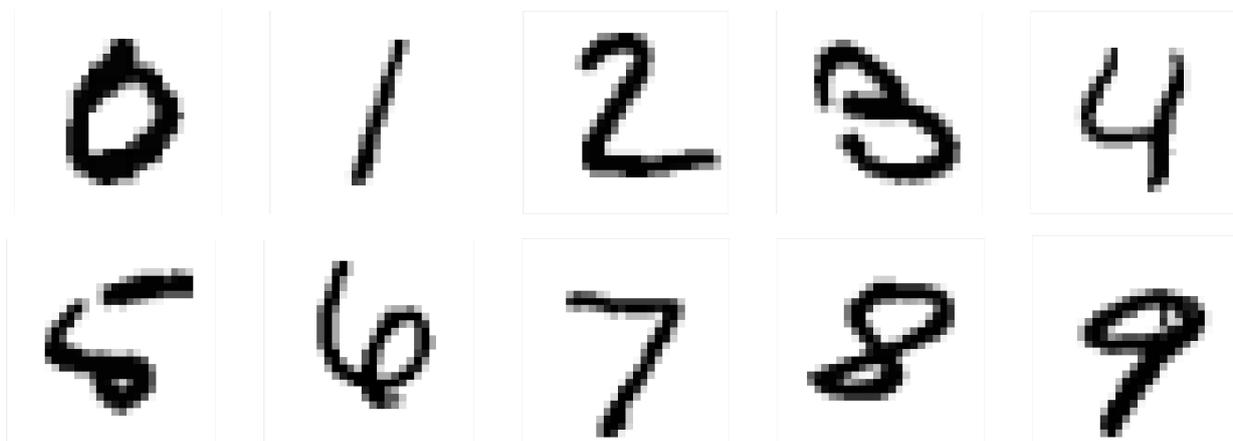

Figure 4 Some data samples from the MNIST dataset

Handwritten digit clustering refers to the problem of clustering multiple handwritten digits with different rotations and translations. We use the handwritten digits from the USPS dataset and MNIST dataset to evaluate the performance of different subspace clustering algorithms. Some data samples from the USPS dataset and MNIST dataset are shown in Fig.3 and Fig.4 respectively. The handwritten digits are hard to cluster for the following reasons. Frist, unlike face images, the images of handwritten digits are less aligned. Second, there are relatively less information contained in each image because each image can be seen as a binary image. Third, as can be seen from the figures, the distances between the subspaces are close to each other because some numbers are similar to others. The above reasons make it difficult to distinguish between these numbers using subspace clustering algorithms. We take all the 10 classes and each class contains 10 samples as the data matrix for subspace clustering. For each database, we take numbers from the training samples and testing samples respectively and they are referred to as train group and test group. The clustering accuracy for the USPS dataset and MNIST dataset is presented in Table 4 and Table 5 respectively.

Table 4 Clustering accuracy (%) of different algorithms on the USPS dataset

| Algorithm | Train group | Test group |
| --- | --- | --- |
| LSA [24] | 63.00 | 69.00 |
| SLBF [25] | 52.00 | 53.00 |
| LLMC [27] | 32.00 | 27.00 |
| SCC [31] | 37.00 | 38.00 |
| MSL [16] | 65.00 | 65.00 |
| SSC [38] | 34.00 | 34.00 |
| LRR [39] | 75.00 | 56.00 |
| LRSC ($P_5$) [40] | 62.00 | 65.00 |
| LRSC ($P_6$) [40] | 62.00 | 73.00 |
| GLLRSC ($P_5$) | 72.00 | 64.00 |
| GLLRSC ($P_6$) | 70.00 | 65.00 |

Table 5 Clustering accuracy (%) of different algorithms on the MNIST dataset

| Algorithm | Train group | Test group |
|---|---|---|
| LSA [24] | 55.00 | 58.00 |
| SLBF [25] | 54.00 | 42.00 |
| LLMC [27] | 33.00 | 47.00 |
| SCC [31] | 33.00 | 28.00 |
| MSL [16] | 61.00 | 60.00 |
| SSC [38] | 60.00 | 57.00 |
| LRR [39] | 24.00 | 36.00 |
| LRSC ($P_5$) [40] | 61.00 | 57.00 |
| LRSC ($P_6$) [40] | 60.00 | 64.00 |
| GLLRSC ($P_5$) | 64.00 | 63.00 |
| GLLRSC ($P_6$) | 66.00 | 64.00 |

From the tables we make the following conclusions. First, we can see that the clustering accuracy of different algorithms is relatively low compared with the clustering accuracy on the Extended Yale B dataset. Second, the clustering accuracy obtained by LRSC is higher than that of LRR which means in this case using a clean dictionary can get higher clustering accuracy and the dictionary learned is closer to a clean dictionary. Third, incorporating manifold information of the data structure can improve the clustering accuracy to some extent. In clustering handwritten digits, using hard thresholding is just as well as using polynomial thresholding. Finally, in order to improve the clustering accuracy on the handwritten digit dataset, some preprocessing should be used. The numbers should be aligned before clustering and aligned images are more appropriate for the subspace clustering models such as LRR and LRSC.

## 5. Conclusions and future research directions

In this paper an improved low rank subspace clustering model is proposed. Manifold information of the data structure is incorporated using the graph Laplacian in the original LRSC model and the resulting model is called graph Laplacian regularized LRSC model. The GLLRSC model is solved by using ADMM optimization method. Experiments on real data set such as face images and handwritten digits show the effectiveness of our algorithm and its superiority over the state-of-the-art.

Here we also point out several possible future research directions. There are many parameters to be determined in the original LRSC model and the proposed GLLRSC model. In this paper, the parameters are chosen heuristically. Finding a method to choose the optimal parameters for these models will be researched in the future. Incorporating manifold information of the data structure is effective for improving the clustering accuracy of various subspace clustering models. Since the manifold information of the data structure can also be described by other methods such as locally linear embedding method, manifold information can also be incorporated by using locally linear embedding into the existing subspace clustering models. In the original LRSC model, a clean dictionary is learned from the data matrix. The learned dictionary may still contain some noises and outliers. In the future research, a more reasonable criterion should be proposed for evaluating the cleanness of the dictionary in order to get higher clustering accuracy. Finally, the original SSC model can also be improved by first learning a clean dictionary and then represent the data samples using this clean dictionary so higher clustering accuracy may be obtained.

*Acknowledgment*

This work is partially supported by the Key Laboratory of Port, Waterway and Sedimentation Engineering of the Ministry of Transport, Nanjing Hydraulic Research Institute; State Key Laboratory of Urban Water Resource and Environment under Grant ES201409; and Key Laboratory of Rivers and Lakes Governance and Flood Protection of Yangtse River Water Conservancy Committee under Grant CKWV2013225/KY.